\documentclass[runningheads]{llncs}
\usepackage{graphicx}

\usepackage{subfig}
\usepackage{amsmath,amssymb} 
\usepackage{color}
\usepackage{cite}

\begin{document}
	
	\title{Gated Transfer Network for Transfer Learning} 
	\titlerunning{Gated Transfer Network} 
	
	
	\author{Yi Zhu\inst{1} \and
		Jia Xue\inst{2} \and
		Shawn Newsam\inst{1}}
	%
	
	\authorrunning{Y. Zhu et al.} 
	
	
	\institute{University of California at Merced, Merced CA 95343, USA 
		\email{\{yzhu25,snewsam\}@ucmerced.edu}\\
		\and
		Rutgers University, New Brunswick NJ 08901, USA\\
		\email{jia.xue@rutgers.edu}
		\footnote{The first two authors contributed equally.}}
	
	\maketitle
	
	\begin{abstract}
		Deep neural networks have led to a series of breakthroughs in computer vision given sufficient annotated training datasets. For novel tasks with limited labeled data, the prevalent approach is to transfer the knowledge learned in the pre-trained models to the new tasks by fine-tuning. Classic model fine-tuning utilizes the fact that well trained neural networks appear to learn cross domain features. These features are treated equally during transfer learning. In this paper, we explore the impact of feature selection in model fine-tuning by introducing a transfer module, which assigns weights to features extracted from pre-trained models. The proposed transfer module proves the importance of feature selection for transferring models from source to target domains. It is shown to significantly improve upon fine-tuning results with only marginal extra computational cost. We also incorporate an auxiliary classifier as an extra regularizer to avoid over-fitting. Finally, we build a Gated Transfer Network (GTN) based on our transfer module and achieve state-of-the-art results on six different tasks.
		
	\end{abstract}
	\section{Introduction}
	\label{sec:intro}
	Deep convolutional neural networks (CNNs) are good at accurately mapping inputs to outputs from large amounts of labeled data. For example, deep residual learning \cite{resnet} already achieves superhuman performance at recognizing objects on the ImageNet challenge \cite{imagenet_cvpr09}. 
	However, these successful models rely on large amounts of annotated data to achieve such performances. For tasks and domains that do not have sufficient training data, deep models may have inferior performance to traditional classification algorithms that utilize hand crafted features. Transfer learning is helpful to deal with these novel scenarios that go beyond tasks where labeled data is abundant. The essence is to transfer the knowledge learned in the pre-trained models to  new domains. 
	
	Transfer learning is a broad research topic, and is studied in several forms such as domain adaptation, multi-task learning and model fine-tuning. 
	In this work, we focus on model fine-tuning and try to answer the question of how to best adapt a pre-trained CNN model to new tasks. 
	We start with the observations of existing model fine-tuning approaches for classification: The common approach is to pre-train a CNN architecture on a source domain with sufficient training data (e.g., ImageNet, which contains 1.2 million images in 1000 categories), and then employ the pre-trained CNN as a feature extraction module, combined with a classification module for the task domain. When the task domain contains limited training data (e.g., CUB-200 \cite{wah2011caltech}, which contains 11 thousand images in 200 categories), two types of prevalent classifier models are employed in transfer learning. The first one is to remove the classification layer from the pre-trained CNN, treat the remaining CNN as a fixed feature extractor, and train a linear classifier for classification. The second type is to replace the pre-trained classification layer with a new classification layer (e.g., a new 200-way classification layer for CUB-200 classification), fine-tuning all (or a portion of) the layers of the network by continuing the back-propagation with a relatively small learning rate. 
	For both scenarios, the underlying thesis is that the well trained CNNs appear to learn cross domain features.
	
	However, both scenarios have their own drawbacks. When we regard a pre-trained CNN as a fixed feature extractor, the features are not tailored for the target domain, which limits their performance. When we fine-tune the pre-trained CNN to learn task specific features in an end-to-end manner, we will encounter the over-fitting problem quite often. How can we best leverage the information from both the source and target domains?
	
	A recent study \cite{cui_FGCV_cvpr2018} shows that there exists a similarity ratio between the source and target domains, termed domain similarity. Inspired by this, we argue that features extracted from pre-trained CNNs also have a similarity ratio: some features share more similarity between the source and target domains than other features. The features with a higher similarity ratio should get more attention. Hence, a feature selection mechanism is expected to be helpful during model fine-tuning. In this paper, we consider some novel questions. \textit{Should the features extracted from the pre-trained CNN be treated equally? Are some features more important and thus be weighted more than others in the novel target domain? Will a feature selection mechanism be helpful for better knowledge transfer?} We aim to answer these questions by introducing a gating architecture, acting like a weight assignment, to the extracted features before classification. We refer to it as a \textit{transfer module} and illustrate the details in later sections.
	Specifically, our contributions include the following:
	\begin{itemize}
		\item First, we prove that a feature selection mechanism is helpful during model fine-tuning and introduce the transfer module as a gating mechanism to decide feature importance when transferring from pre-trained CNNs to target domains.
		\item Second, we incorporate an auxiliary classifier to stabilize the training and avoid over-fitting.
		\item Finally, our proposed gated transfer network achieves state-of-the-art results on six different domains, which demonstrates its superior transferability.
	\end{itemize}

	\section{Related Work}
	\label{sec:related}
	
	There is a large body of literature on transfer learning; here we review only the most related work. 
	Starting from \cite{gong2014multi}, off-the-shelf CNN features have shown promising performance in various computer vision domains. 
	Later, Yosinski et al. \cite{Yosinski_transfer_nips2014} raised the question of ``How transferable are features in deep neural networks'', and experimentally quantified the generality versus specificity of neurons in each layer of a deep CNN. However, features from the pre-trained models are not tailored for the target domain, which limits their performance. The common practice now is to not only replace and retrain the classifier on top of the CNN on the new dataset, but also fine-tune the weights of the pre-trained network by continuing the back-propagation, which is known as model fine-tuning \cite{Oquab_midlevel_cvpr2014}. Depending on the size of the new dataset, we could either train all the layers or only part of the network. Hence, task-specific features could be learned in an end-to-end manner. 
	
	Recently, there have been several attempts  to explore more information between the source and target domains for better knowledge transfer \cite{Li_learn_wo_forget_eccv2016,Liu_sparse_transfer_aaai2017}. \cite{grow_brain_cvpr2017} tries to grow a CNN with additional units, either by widening existing layers or deepening the overall network, and achieves moderate success over traditional fine-tuning. \cite{ge_wealthy_cvpr2017} proposes a source-target selective joint fine-tuning scheme based on the observation that low-level characteristics (shallow network features) are similar between the source and target domains. Their method obtains state-of-the-art results on several benchmarks. However, the joint fine-tuning pipeline consists of multiple stages and is not end-to-end optimized. \cite{cui_FGCV_cvpr2018} introduces a measure to estimate domain similarity via the Earth Mover's Distance and demonstrates that transfer learning benefits from pre-training on a source domain that is similar to the target domain by such a measure. 
	
	Our work lies in the recent direction that explores the source-target domain relationship, and is most similar to \cite{grow_brain_cvpr2017} in terms of increasing model capacity for more natural model adaptation through fine-tuning. However, we differ in several aspects.  
	First, \cite{grow_brain_cvpr2017} makes an analogy to developmental learning while our work is inspired by sparse learning. Our transfer module acts like a gating mechanism without learning new features. We are interested in how to make existing features more distinctive for various target domains. Thus, we can better handle over-fitting in the insufficient training data scenario.
	Second, \cite{grow_brain_cvpr2017} proposes to make the fully connected layers deeper or wider to increase model capacity. 
	This technique does not scale well to recent deeper networks like ResNet \cite{resnet} or DenseNet \cite{densenet} because they do not have fully connected layers. In contrast, our proposed transfer module is generally applicable. 
	Third, we design our transfer module based on a ``Squeeze-and-Excitation'' block \cite{senet_cvpr2018}. Squeeze aims at embedding global information while excitation learns to use the global information to selectively emphasize informative features and suppress less useful ones. We show that our transfer module outperforms \cite{grow_brain_cvpr2017} on several widely adopted benchmarks for transfer learning. 
	

	\begin{figure*}[t]
		\centering
		\includegraphics[width= 1.0\linewidth]{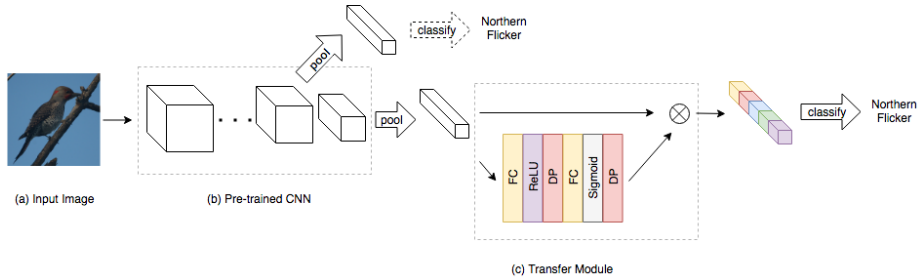}
		\caption{Overview of our proposed gated transfer network. Our idea is that feature selection makes more sense than learning new features when we fine-tune a pre-trained CNN. The transfer module functions as a gating mechanism to assign weights to features from the pre-trained CNN. $\otimes$ denotes the point-wise multiplication. An auxiliary classifier is only used during the training phase. }
		\label{fig:structure}
	\end{figure*}

	\section{Gated Transfer Network}
	
	In this section, we first introduce our transfer module as a feature attention mechanism and interpret its relation to other approaches. Then we incorporate the auxiliary classifier to stabilize the model fine-tuning and avoid over-fitting to domains with insufficient training data. Finally, we describe the details of our proposed gated transfer network.
	
	\subsection{Transfer Module}
	A weight assigning mechanism, which transforms features from the source domain to target domain, is very important for transfer learning. Intuitively, knowledge transfer will be easier if the source and target domains are similar: models trained on more images will likely to learn features in an unbiased manner, and lead to better transfer learning performance. However, as studied in \cite{sun_JFT_300M_iccv2017,Miao_2018_CVPR}, transfer learning performance is logarithmically related with the scale of training data. This indicates that the performance gained from using more training data would be insignificant when we already have a large-scale dataset (e.g., ImageNet \cite{imagenet_cvpr09} or Places \cite{zhou_places_nips2014}). Hence, feature selection makes more sense than learning new features when we fine-tune a pre-trained CNN. 
	
	The ground rule of our approach is that there should be common features (knowledge) between the source and target domains in order to do transfer learning. We should give more attention to the common features and ignore the dissimilar ones. In addition, during back propagation, the common features should be focused on and finely adjusted, instead of treating all features from a pre-trained model equally.
	
	Let $x$ denote the extracted features from the aforementioned feature extractor module, with $y$ denoting the target features. Our transfer module is defined as:
	\begin{equation}
		y = F(x) \otimes x
	\end{equation}
	where the function $F(x)$ is the gating function and represents the feature map weights to be learned. $\otimes$ is the channel-wise multiplication between the feature maps $x$ and the feature weights $F(x)$. When $F(x)$ contains all 1s, it passes all the learned features through without gating, which is basically a classic fine-tuning approach. We will provide more interpretations on our proposed transfer module in the next section. 
	
	In addition, over-fitting is a non-trivial problem in transfer learning, especially when the training images are limited. 
	To prevent over-fitting, we adopt the dropout\cite{srivastava2014dropout} function in our transfer module. 
	As shown in Figure~\ref{fig:structure}, our transfer module is designed as: 
	\begin{equation}
		F(x) = \rho \delta ({W}_{2} \rho \sigma ({W}_{1}x))
	\end{equation}
	in which $\sigma$ denotes the ReLU function, $\rho$ refers to the dropout function, and $\delta$ is the sigmoid function. ${W}_{1}$ and ${W}_{2}$ are two fully-connected layers. 
	
	\subsection{Interpretation}
	We introduce the transfer module as a feature selection mechanism into CNN fine-tuning approaches with minimal extra computational cost. Here, we interpret why it works by relating it to classic fine-tuning and sparse coding. 
	
	\paragraph{\bf Classic Fine-tuning} 
	Classic fine-tuning replaces the pre-trained classification layer with a new classification layer for the target domain. All (or a portion of) the layers of the network continue back-propagation until the new model converges. It does not consider feature importance during the knowledge transfer process. If our transfer module learns the extracted features as being equally important for the target domain, our network will assign same weights to the features. In the end, our transfer module will simply become the classic fine-tuning approach. 
	
	\paragraph{\bf Sparse Coding} 
	Sparse coding assumes that a task's parameters are well approximated by sparse linear combinations of the atoms of a dictionary on a high or infinite dimensional space. The sparsity acts as regularization and works well for transfer learning tasks. When considering sparse coding in CNNs, most literature utilizes it for model compression and reduced computational cost. The overall idea is that some sparse representations embedded in the pre-trained model are vital for visual recognition. 
	If our transfer module learns that only a few features are important for the target domain, our network will assign high weights to these important features and low weights to the other features. In the end, our transfer module will function like sparse coding.
	
	\subsection{Auxiliary Classifier}
	
	\cite{vgg1619,resnet,inception_v1_cvpr2015} demonstrate that deeper networks generally lead to better performance in images classification. However, a very deep network may be unstable during training, due to vanishing gradients and information forgetting \cite{resnet}. In the transfer learning regime, the problems are even more severe because of the additional optimization difficulty brought by the limited training set. 
	
	Motivated by the observation in \cite{inception_v1_cvpr2015}, we believe that features produced by the intermediate layers of the network should be very discriminative. Instead of only using the output layer, we can provide additional feedback signals to the network by adding auxiliary classifiers after the intermediate representations. In this manner, we expect to learn more distinctive features for the target domain in the lower layers of the network. Furthermore, the extra gradient signal will provide additional regularization, and alleviate the common problem of vanishing gradients. 
	
	Hence, we introduce an auxiliary classifier as shown in Figure \ref{fig:structure}. The total loss of the network is the sum of original loss from the output layer and the new loss from the auxiliary classifier. We aim to let the auxiliary classifier and transfer module separately do their own jobs. The auxiliary classier learns realizable features for the target domain, and the transfer module selects the features based on the attention weights. We only employ the auxiliary classifier during the training phase, and in later sections, we will show that the auxiliary classifier provides performance improvement even when there are already multiple dropout layers and batch normalization. Thus, we have the same observation as in \cite{inception_v2_cvpr2016} that the auxiliary classifier acts as regularizer.

	\subsection{Network Architecture}
	
	With the proposed transfer module and auxiliary classifier, we build our gated transfer network as illustrated in Figure~\ref{fig:structure}. Given an input image, we use a pre-trained CNN while removing the last fully connected layer to extract the feature maps. The pre-trained CNN could be of any architecture, e.g., AlexNet, VGGNet, ResNet, DenseNet, etc. This is followed by the transfer module to assign attention weights to the features. The auxiliary classifier is on another branch to provide feedback for more regularization. In our experiments, we branch it out before the last convolutional group. Finally, the network makes a prediction based on the gated features with a new classification layer for the target domain. 
	
	To improve model efficiency and aid generalization, inspired by \cite{resnet,senet_cvpr2018}, we reduce the feature map channels between the fully connected layers in the transfer module with a manually defined reduction ratio $r$. If the channel for input features is $C$, then the dimension of the weights for the first fully connected layer is ${W}_{1}\in {\mathbb{R}}^{\frac{C}{r}\times C}$ and the dimension of the weights for the second fully connected layer is ${W}_{2}\in {\mathbb{R}}^{C\times \frac{C}{r}}$. In our experiments, we set the reduction ratio to 16 for a good trade-off between accuracy and efficiency.

	\section{Experiments}
	
	\subsection{Datasets}
	We choose six different datasets for evaluating of our proposed transfer module: MIT67, CUB200, MINC, DTD, UCF101 and FashionAI. MIT67 and CUB200 are widely adopted benchmarks for evaluating CNN transferability. We also report performance on four other large-scale datasets, MINC for material recognition, DTD for texture classification, UCF101 for video human action recognition and FashionAI for attribute recognition of apparel. These tasks are quite challenging for existing transfer learning techniques because they are conceptually different from what the pre-trained CNN was designed to do (e.g., object recognition). We follow the standard experimental setup (e.g., the train/test splits) for these datasets, and demonstrate the effectiveness of our approach in later sections.
	
	In brief, MIT Indoor 67 (MIT67) \cite{mit67} is an indoor scene categorization dataset with 67 categories. A standard subset of 80 images per-category for training and 20 for testing is used. 
	Caltech-UCSD Birds 200-2011 (CUB200) \cite{wah2011caltech} is a fine grained recognition dataset containing 200 categories of birds. There are 5994 training images and 5794 test images. 
	For MIT67 and CUB200, we report the mean class accuracy on the test set. 
	Materials in Context Database (MINC) \cite{bell15minc} is a large scale material dataset. In this work, a publicly available subset MINC-2500 is evaluated, containing 23 material categories and 2,500 images per-category. 
	Describable Textures Dataset (DTD) \cite{cimpoi14describing} is a collection of textural images in the wild. 
	This texture database consists of 5640 images and is organized according to a list of 47 categories. 
	UCF101 \cite{ucf101} is composed of realistic action videos from YouTube. It contains 13320 video clips distributed among 101 action classes. Each video clip lasts for an average of 7 seconds. 
	For MINC, DTD and UCF101, we report the average recognition accuracy. 
	FashionAI \cite{fashionai} is a global challenge aiming to make AI insights on fashion. There are two tasks in the challenge, and we focus on the attribute recognition of apparel task. The training dataset has 79573 images from 8 attributes. 
	Each attribute has multiple attribute values. 
	We submit our results on the held out test set to the evaluation server, and report the mean average precision. 
	
	\subsection{Implementation Details} 
	
	We use stochastic gradient descent (SGD) with a mini-batch size of 64 for our deep model training and evaluation. For data augmentation, the input images are first resized to 256 along the short edge, and normalized by pixel mean subtraction and standard deviation division. The images are then randomly cropped to size 224, keeping the aspect ratio between 3/4 and 4/3. We also flip the images horizontally with a 50$\%$ chance. For fine-tuning, the learning rate starts at 0.01 and is divided by 10 when the validation error plateaus. During fine-tuning, we first freeze the pre-trained CNN for the first several epochs to get a good initialization of our transfer module, and then we make them learn at the same pace. We use a weight decay of 0.0001 and a momentum of 0.9. In testing, we adopt the standard center crop evaluation strategy.
	
	\subsection{Ablation Study}
	In this section, we will perform a series of ablation studies on MIT67, MINC and CUB200, to both justify our network design and demonstrate the effectiveness of our proposed transfer module. To be specific, we first show how significant dropout can help the fine-tuning process to avoid over-fitting. Second, we show the impact of the auxiliary classifier through a grid search of loss weights. Third, we compare with a strong baseline method \cite{grow_brain_cvpr2017} to indicate the superiority of our gating mechanism. Fourth, we compare our transfer module with widely adopted residual block \cite{resnet} to confirm our design choice. 
	For all the experiments, we use ResNet50 as the backbone CNN architecture.
	
	\begin{table}[t]
		\begin{center}
			\caption{Impact of dropout. TM: transfer module. AUX: auxiliary objective. Numbers in parentheses are mean accuracies with dropout. \label{tab:dropout}}
			
			\begin{tabular}{ c | c | c | c }
				\hline
				Method			         &    MIT67    &    CUB200      &    MINC         \\
				\hline		
				\hline
				Baseline fine-tuning   	 &   $76.95$  &   $77.64$ 	    &    $77.88$   	\\
				\hline
				Pre-trained + TM   &   $77.17$ ($77.29$)  &   $77.85$  ($79.72$)	   &    $79.83$  ($80.41$) \\
				\hline
				Pre-trained + TM + AUX	 &   $77.54$  ($\mathbf{77.77}$) &   $77.96$  ($\mathbf{79.95}$)	    &    $\mathbf{81.05}$  ($80.80$)  	\\
				\hline
			\end{tabular}
			
		\end{center}
		
	\end{table} 
	
	\paragraph{\bf Dropout as Good Practice:}
	For transfer learning to a target domain with little annotated data, over-fitting is usually the most challenging problem to deal with. This is true for classic fine-tuning as well as more recent domain adaption approaches, and our approach is no exception.
	
	As shown in Table \ref{tab:dropout}, our transfer module brings significant improvement over baseline fine-tuning on all three benchmarks. Here, baseline indicates directly fine-tuning on the target domain using a ResNet50 model pre-trained on ImageNet. Similarly, due to the extra regularization, our model with both the transfer module and auxiliary objective performs the best. However, we observe that for CUB200, the improvement is rather limited (77.64 $\Rightarrow$ 77.96). On the other hand, our proposed method achieves much higher accuracy (77.88 $\Rightarrow$ 81.05) than the baseline on the MINC dataset. We believe that we are experiencing over-fitting since CUB200 has too little training data (e.g., 20 images per class), while MINC has many more images (e.g., 2500 images per class).  
	
	Hence, we incorporate dropout \cite{srivastava2014dropout} as a good practice in our transfer module. As shown in Figure \ref{fig:structure}, we add two dropout layers after each fully connected layer, with dropout rates of 0.5 and 0.7, respectively. The dropout rates are empirically determined based on cross-validation. The numbers inside the parentheses in Table \ref{tab:dropout} are the recognition accuracy after adding dropout layers. We observe a clear performance boost on CUB200 (77.64 $\Rightarrow$ 79.95), and moderate improvement on MIT67 and MINC. This confirms that our initial model without dropout layers suffers from serious over-fitting on CUB200. There is one interesting observation that adding dropout does not improve MINC (81.05 $\Rightarrow$ 80.80) when the auxiliary classifier is also present. This is because the auxiliary loss can regularize the training given sufficient data, and thus there is no need for extra dropout layers. Next, we will investigate the impact of the auxiliary loss.
	
	\paragraph{\bf Impact of Auxiliary Classifier:} 
	Intermediate features are also capable of discriminating different classes because they are extracted from earlier layers in which the gradients carry information. Hence, the auxiliary classifier is introduced to regularize the network fine-tuning especially when the training data is small. Here, we perform an ablation study to justify its impact. 
	
	As show in Table \ref{tab:aux}, we assign different weights to the auxiliary loss during CNN optimization. A loss weight of 0 indicates we do not use the auxiliary loss. We can see that the auxiliary loss does bring improvement on all three datasets. The optimal weight of the auxiliary loss dependa on the dataset. We choose $0.2$ as a trade-off, and use it for all our experiments hereafter. 
	Although the improvements seem marginal, they are actually significant considering we already use state-of-the-art techniques such as batch normalization and dropout to regularize model training. In addition, our model converges about $10\%$ faster due to the strong supervision signals flowing to earlier layers in the network. 
	
	\begin{table}[t]
		\begin{center}
			\caption{The impact of the weight of the auxiliary loss on classification performance. We find that the auxiliary loss performs best with a weight of 0.2 and use this value for the remaining experiments. \label{tab:aux}}
			
			\begin{tabular}{ c | c | c | c }
				\hline
				Loss weights			         &    MIT67    &    CUB200      &    MINC         \\
				\hline		
				\hline
				0    	 &   $77.29$  &   $79.72$ 	    &    $80.41$   	\\
				\hline
				0.1    	 &   $77.09$  &   $78.93$ 	    &    $80.56$   	\\
				\hline
				0.2   &   $\mathbf{77.77}$   &   $\mathbf{79.95}$     &    $80.80$  \\
				\hline
				0.4	 &   $77.07$   &   $79.81$  	    &    $80.47$  	\\
				\hline
				0.8	 &   $76.27$   &   $79.24$  	    &    $\mathbf{80.87}$  	\\
				\hline
			\end{tabular}
			
		\end{center}
		
	\end{table}

	\paragraph{\bf Transfer Module vs Grow a Brain \cite{grow_brain_cvpr2017}:} 
	We mentioned in Section \ref{sec:related} that \cite{grow_brain_cvpr2017} is the most related work. \cite{grow_brain_cvpr2017} proposed several structures, such as depth-augmented (DA)-CNN, width-augmented (WA)-CNN, etc., but the basic idea is to deepen or widen the fully connected layers in the CNN model. Here, we compare our GTN to this strong baseline \cite{grow_brain_cvpr2017}. 
	
	Because \cite{grow_brain_cvpr2017} reported its performance using VGG16, for fair comparison, we report our numbers using both VGG16 and ResNet50. As we can see in Table \ref{tab:grow}, our recognition accuracies outperform \cite{grow_brain_cvpr2017} by a large margin on MIT67 and CUB200 using VGG16. 
	We also re-implement \cite{grow_brain_cvpr2017} using ResNet50. Since ResNets do not have fully connected layers, we could not width-augment the network. Thus, to depth-augment the network as in \cite{grow_brain_cvpr2017}, we add one fully connected layer and one batch normalization layer before the classifier. As shown in in Table \ref{tab:grow}, our GTN with ResNet50 still obtains better performance than \cite{grow_brain_cvpr2017} with ResNet50 on all three benchmarks. 
	Note that, for MIT67, the depth-augmented ResNet50 performs even worse than the baseline model fine-tuning (as shown in Table \ref{tab:dropout}, 75.38 $<$ 76.95). This indicates that \cite{grow_brain_cvpr2017} may not generalize well to deeper architectures like ResNet and DenseNet.
	
	\begin{table}[t]
		
		\begin{center}
			\caption{Comparison with DA-CNN\cite{grow_brain_cvpr2017} using both VGG16 and ResNet50 models. GTN outperforms DA-CNN on all three datasets. \label{tab:grow}}
			
			\begin{tabular}{ c | c | c | c }
				\hline
				Method			         &    MIT67    &    CUB200      &    MINC         \\
				\hline		
				\hline
				\cite{grow_brain_cvpr2017} with VGG16 	 &   $66.3$  &   $69.0$ 	    &    $-$   	\\
				\hline
				GTN with VGG16    	 &   $71.19$  &   $72.14$ 	    &    $-$   	\\
				\hline
				\cite{grow_brain_cvpr2017} with ResNet50    	 &   $75.38$  &   $76.28$ 	    &    $78.56$   	\\
				\hline
				GTN with ResNet50   &   $\mathbf{77.77}$  &   $\mathbf{79.95}$     &    $\mathbf{80.80}$   \\
				\hline
			\end{tabular}
			
		\end{center}
	\end{table}

	\begin{table}[h]
		
		\begin{center}
			\caption{Performance comparisons of classification accuracy (\%) on the target datasets between the transfer module (multiplication) and a residual block (summation) based on ResNet50.  \label{tab:residual}}
			
			\begin{tabular}{ c | c | c | c }
				\hline
				Method			         &    MIT67    &    CUB200      &    MINC         \\
				\hline		
				\hline
				Summation   	 &   $76.47$  &   $78.39$ 	    &    $80.23$   	\\
				\hline
				Multiplication  &   $\mathbf{77.29}$  &   $\mathbf{79.72}$     &    $\mathbf{80.41}$   \\
				\hline
			\end{tabular}
			\label{tab:residual}
		\end{center}
		
	\end{table}
	
	\paragraph{\bf Transfer Module vs Residual Block:} 
	
	Continuing the discussion from the last paragraph, it is straightforward to grow network capacity by deepening or widening fully connected layers in AlexNet and VGG16. However, for recent widely-adopted deeper networks like ResNet and DenseNet, it is more natural to add residual blocks to grow the network capacity. A residual block \cite{resnet} aims to utilize additional shortcut connections to pass the input directly to the output and ease the model optimization procedure. But it still needs to learn the corresponding residuals. In this work, we argue that features from models pre-trained on ImageNet are already good enough. We only need to select those features instead of learning new ones. 
	
	Simply, if we design our transfer module by replacing the last multiplication with a summation operation in Figure \ref{fig:structure}, we will transform it to a standard residual block. In this way, we expect the network to learn residuals to make the feature set more distinctive. The comparison results are shown in Table \ref{tab:residual}. We observe that multiplication obtains better results than summation, especially on CUB200 (78.39 $\Rightarrow$ 79.72). This indicates that our transfer module with multiplication is more easily optimized when dealing with insufficient data. The reason is because we only need to learn how to select the original features instead of learning something new. 
	
	\begin{figure}[t]
		\centering
		\includegraphics[width=1.0\linewidth]{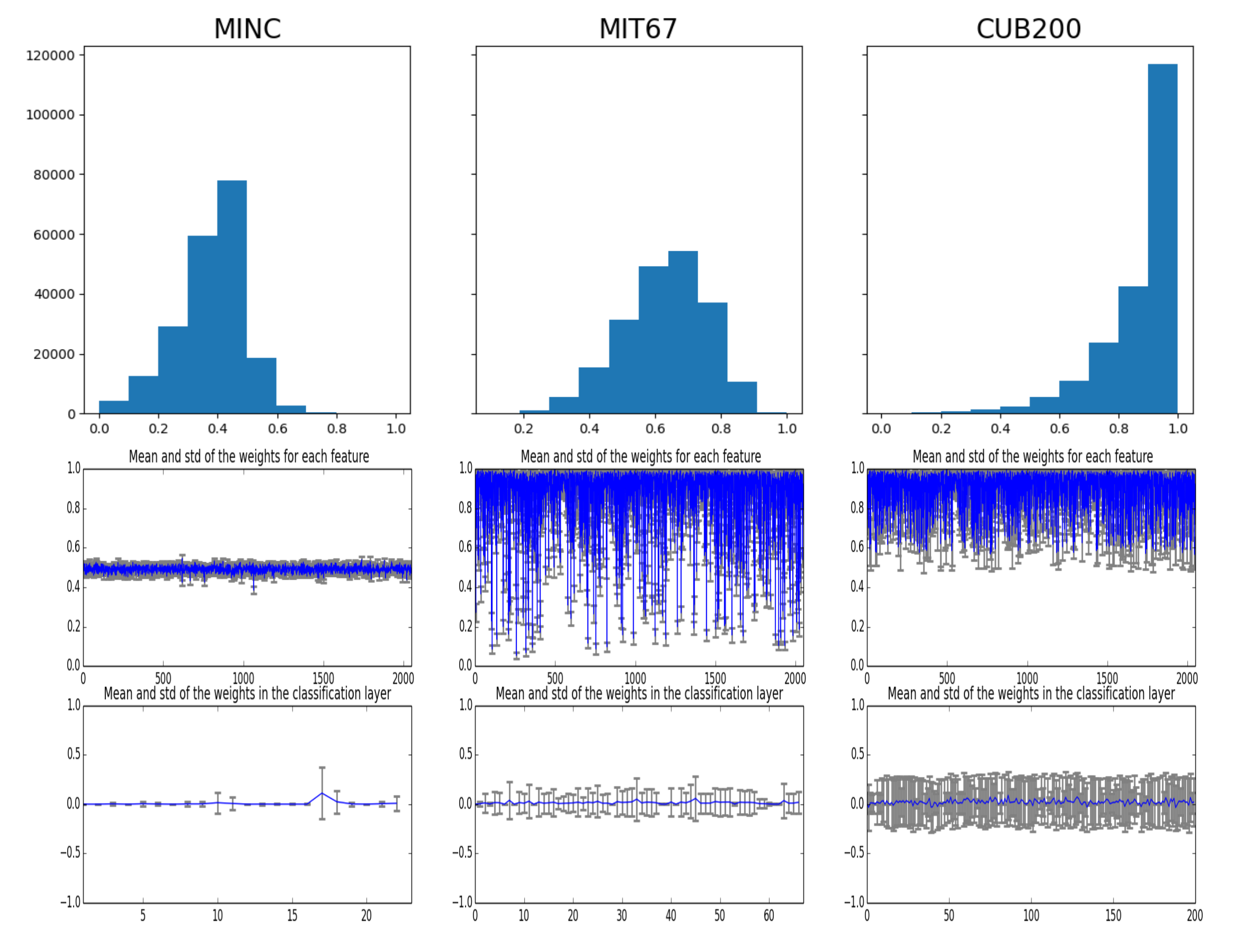}
		
		\caption{\textbf{Top}: Histogram of the attention weights from the transfer module. It visualizes how our transfer module selects useful features for the target domain. Values on the Y axis indicate the number of neurons. Our transfer module indeed functions as a gating mechanism based on the domain similarity. \textbf{Bottom}: Mean (blue) and standard deviation (gray) of the weights for each feature in our gating module and in the classification layer. Same observation holds. }
		\label{fig:vis}
	\end{figure}
	
	\begin{figure}[t]
		\centering
		\subfloat
		{
			\includegraphics[width=.3\linewidth]{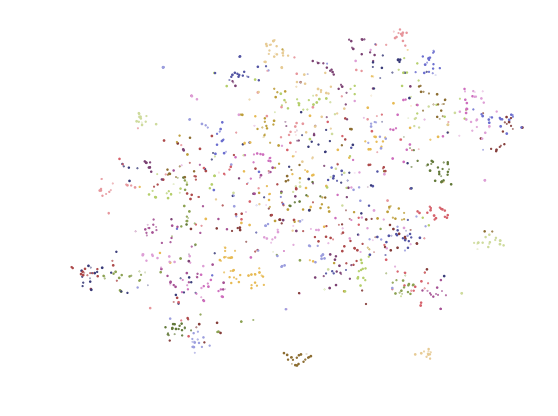}
		}
		\subfloat
		{
			\includegraphics[width=.3\linewidth]{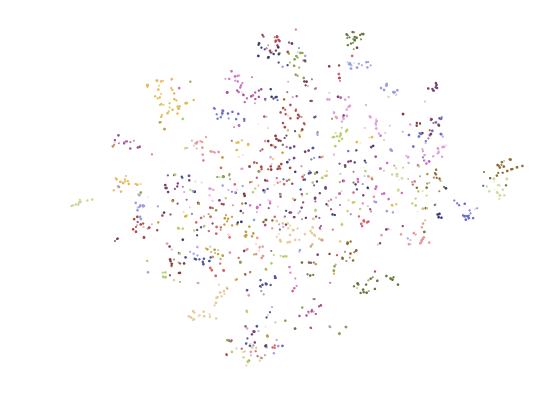}
		}
		\subfloat
		{
			\includegraphics[width=.3\linewidth]{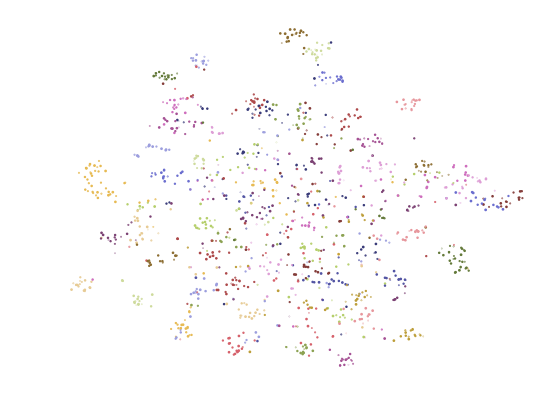}
		}
		\setcounter{subfigure}{0}
		\subfloat[Pre-Trained Network]
		{
			\includegraphics[width=.3\linewidth]{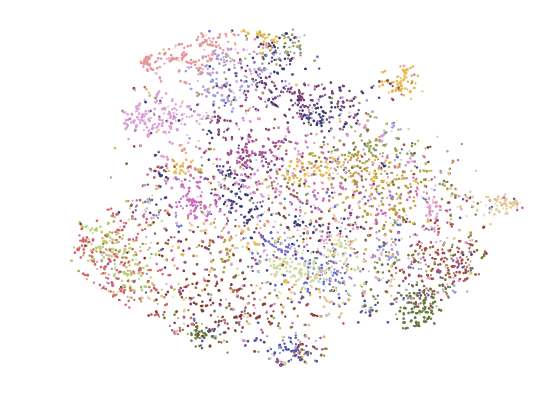}
		}
		\subfloat[DA-CNN\cite{grow_brain_cvpr2017}]
		{
			\includegraphics[width=.3\linewidth]{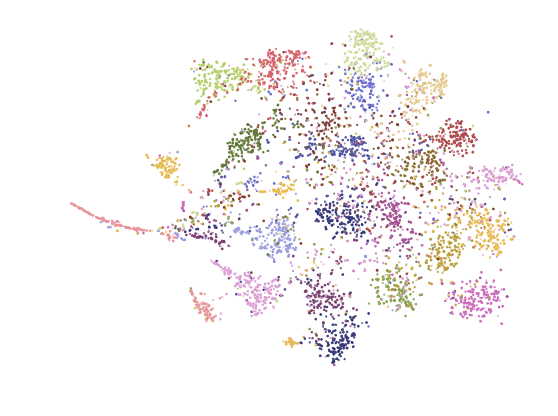}
		}
		\subfloat[Gated Transfer Network]
		{
			\includegraphics[width=.3\linewidth]{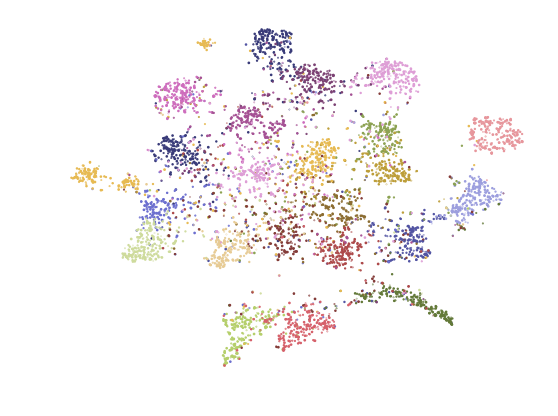}
		}
		\caption{t-SNE visualization of the features from top layers on the MIT67 (top) and MINC (below) validation set. Our GTN shows better semantic separations.}
		\label{fig:tsne}
	\end{figure}

	\subsection{Visualizing Transfer Module}
	
	To demonstrate what the transfer module is learning, we visualize its weight distribution. Since the last layer of the transfer module is a sigmoid activation function, the output will be in the range [0, 1]. The dimension of the output is the same as the original features from the pre-trained model (e.g., 2048 in our case). This represents a one-to-one gating score. A value closer to 1 indicates the feature is important for target domain classification, while a value closer to 0 means it is irrelevant.
	
	Here, we show three histograms to visualize the selection weight distribution. For each dataset, we randomly select a batch of 100 images as input. For the x-axis, we divide 0$\sim$1 into 10 bins, and group the values of neurons in each group. As we can see in Figure \ref{fig:vis} top, the selection score distributions are different for different datasets. Such distributions could be used to describe the domain similarity between source and target domains. MINC is a material recognition dataset, which is quite different from ImageNet. It requires low-level information instead of the high-level object features. Hence, most of the weights are small (below 0.5), which means the original features are not important for the target domain. MIT67 is a scene classification dataset and, although different from the object recognition task, most of the scenes contain objects. There are some common features (knowledge) that can be transferred, thus most of the features have a gating score of 0.7. CUB200 is a fine-grained dataset for bird species recognition. It is also an object recognition task like ImageNet. In addition, some of the images are from ImageNet. Hence, the source domain and target domain are very similar. As expected, most weights are large (above 0.7), which means most of the original features (knowledge) are useful for the target domain. 
	The observation demonstrates that our transfer module indeed functions as a gating mechanism. We select useful features based on the domain similarity between the source and target domains.
	
	In addition, we plot the mean and standard deviation of the weights for each feature in the gating module and the weights in the classification layer in Figure \ref{fig:vis} bottom. We have the same observation as demonstrated by the histograms. The higher the domain similarity (CUB200 $>$ MIT67 $>$ MINC) between the source and target domains, the more features we select for the target application. 
	
	To gain insight into the effect of the transfer module in feature space, we also visualize the features extracted from the MIT67 and MINC validation datasets using the t-SNE algorithm \cite{van2014accelerating}. As shown in Figure~\ref{fig:tsne}, based on ResNet50, we embed the 4096 dimension features before the classification layer of the pre-trained networks, DA-CNN and GTN, into a 2 dimensional space and plot them as points colored based on their semantic categories. From the t-SNE manifold, our GTN consistently shows better semantic separation, which is compatible with its improved classification performance.
	
	\subsection{Learning without Forgetting}
	
	A must-have property of transfer learning is that new tasks can be learned without suffering from catastrophic forgetting. In particular, for a CNN model, we would like to use the same or similar sets of parameters to adapt to new target domains without sacrificing accuracy on the source domain. 
	
	Initial attempts include feature extraction and model fine-tuning for which we have already noted the drawbacks in Section \ref{sec:intro}. In order to avoid domain shift, a joint training scheme is proposed. Simply put, all the model parameters are jointly optimized by interleaving samples from all tasks. This method's performance may be considered as an upper bound of a model's robustness.
	However, joint training becomes increasingly cumbersome as more tasks are learned and is not possible if the training data for the previously learned tasks is unavailable. There are quite a few works making progress in this direction \cite{Li_learn_wo_forget_eccv2016,grow_brain_cvpr2017}. Here, we compare to two recent state-of-the-art works that focus on learning without forgetting, \cite{Li_learn_wo_forget_eccv2016} and \cite{grow_brain_cvpr2017}. 
	
	We take our fine-tuned model and re-fine-tune it on the source domain, which is the ImageNet training set \cite{imagenet_cvpr09}. We fix the convolutional layers and only fine tune the transfer module and the new 1000-way classifier. For fair comparison, we report accuracy using an AlexNet architecture. 
	As shown in Table \ref{tab:lwf}, our proposed GTN consistently outperforms previous methods, regardless of which target domain it is fine-tuned on. Even when the target domain is video action recognition (quite different from image object classification), our re-trained model can achieve higher accuracy than LwF \cite{Li_learn_wo_forget_eccv2016} and DA-CNN \cite{grow_brain_cvpr2017}. 
	Oracle indicates the original AlexNet trained on ImageNet. Note that we almost achieve the upper bound set by joint training. We believe such robustness of our approach is because our gating mechanism aims to select features instead of learning new ones, thus we are not designed to ``forget''. In addition, other existing approaches \cite{Li_learn_wo_forget_eccv2016} can be naturally incorporated into our approach to further improve the performance on both source and target tasks.
	
	\begin{table}[t]
		
		\begin{center}
			\caption{Learning without forgetting on the ImageNet \cite{imagenet_cvpr09} validation set. Experiments based on an AlexNet architecture. The dataset in brackets indicates the target domain on which the model is fine-tuned. Our proposed GTN consistently outperforms previous approaches, regardless of which target domain it is fine-tuned on. \label{tab:lwf}}
			
			\begin{tabular}{ c | c }
				\hline
				Method 		 &    Acc ($\%$)           \\
				\hline		
				\hline
				Oracle   	 &   $56.9$  	\\
				\hline
				LwF \cite{Li_learn_wo_forget_eccv2016}   	 &   $55.9$    	\\
				Joint \cite{Li_learn_wo_forget_eccv2016}   	 &   $56.4$    	\\
				DA-CNN   \cite{grow_brain_cvpr2017} 	 &   $55.3$   	\\
				WA-CNN \cite{grow_brain_cvpr2017}  &   $51.5$    \\
				\hline
				GTN (CUB200) &   $56.3$    \\
				GTN (UCF101) &   $56.2$    \\
			\end{tabular}
			\label{tab:residual}
		\end{center}
		
	\end{table}
	
	\subsection{Comparison to State-of-the-art}
	
	We evaluate our proposed transfer module on six benchmarks: MIT67 for scene classification, CUB200 for fine-grained recognition, MINC for material recognition, DTD for texture classification, UCF101 for video action recognition and FashionAI for apparel attribute recognition. Here, we use ResNet152 as the backbone CNN architecture. 
	CUB200 and MIT67 are widely adopted benchmarks to evaluate CNN transferability, thus we compare to previous transfer learning literature. For the other four datasets, we compare to transfer learning approaches as well as to the popular baselines for each dataset.
	
	As shown in Table \ref{tab:sota}, our proposed GTN consistently outperforms previous state-of-the-art results on all benchmarks. 
	Most importantly, our transfer module can be combined with other transfer learning techniques \cite{ge_wealthy_cvpr2017} or deeper networks \cite{densenet} for further improvement. 
	Note that \cite{ge_wealthy_cvpr2017} reported a better result on MIT67 than us. However, they used a better CNN backbone (identity mapping based ResNet152) and extra training data (Places dataset \cite{zhou_places_nips2014}) for the source domain. We believe this is not a fair comparison. Since MIT67 is a scene classification dataset, it has higher domain similarity with Places rather than ImageNet. For equal comparison, we re-implement \cite{ge_wealthy_cvpr2017} using vanilla ResNet152 without the extra training data from Places, and our GTN outperforms it by $0.9\%$. 
	
	\begin{table}[t]
		\begin{center}
			\caption{Performance comparisons of classification accuracy (\%) with previous work on scene classification, fine-grained recognition, material recognition and action recognition. $^{*}$ indicates our reimplementation result. Our approach achieves state-of-the-art  performance  on  these  challenging  benchmark  datasets. \label{tab:sota}}
			
			\begin{tabular}{ c | c || c | c || c | c || c | c || c | c || c | c }
				\hline
				\multicolumn{2}{c}{MIT67}     &   \multicolumn{2}{c}{CUB200}      &    \multicolumn{2}{c}{MINC}   &    \multicolumn{2}{c}{DTD}    &    \multicolumn{2}{c}{UCF101}      &    \multicolumn{2}{c}{FashionAI}        \\
				\hline
				& Acc &  & Acc &  & Acc &  & Acc &  & Acc &  & Acc \\
				\hline
				\cite{Li_learn_wo_forget_eccv2016}   &   $64.5$  &  \cite{Li_learn_wo_forget_eccv2016}   &  $56.6$   & \cite{cimpoi2015deep} & 63.1  &  \cite{cimpoi2015deep} & 72.3 &  \cite{depth2action}	&  $63.3$  &  \cite{resnet}  &	 $84.2$ \\
				\cite{grow_brain_cvpr2017}  &   $66.3$  & \cite{grow_brain_cvpr2017}  &  $57.7$ 	    &  \cite{grow_brain_cvpr2017}  &  $80.4^{*}$  & \cite{grow_brain_cvpr2017}  & $71.7^{*}$  &  \cite{hidden_zhu_18}  & $82.4$	    &  \cite{grow_brain_cvpr2017}  &  $86.5^{*}$	\\
				\cite{ge_wealthy_cvpr2017}    	 &   $78.1$  & \cite{Zhang_part_rcnn_eccv2014}   	    &   $73.9$  &  \cite{zhang2016deep}  &   80.4	    &    	
				\cite{zhang2016deep}      &   69.6	    & \cite{dovf_lan_2017} &  $79.5$ 	    &  \cite{vgg1619} &  $81.2$ 	  \\
				\cite{Liu_sparse_transfer_aaai2017}    	 &   $72.5$  &    \cite{Branson_posenorm_bmvc2014}	    &  $75.7$   &  \cite{xue2018deep}  &  $81.2$  &    \cite{xue2018deep}	  &   73.2    &   \cite{url_cvpr18}     &  $82.3$ 	    &  \cite{inception_v1_cvpr2015}   &  $85.6$ 	    \\
				\hline
				\hline
				GTN  &   $\mathbf{79.0}$  & GTN  &  $\mathbf{83.4}$ 	    & GTN  &   $\mathbf{81.3}$  &  GTN  & $\mathbf{74.3}$  &  GTN  & $\mathbf{85.6}$ 	    &   GTN  & $\mathbf{92.0}$ 	\\
			\end{tabular}
			
		\end{center}
		
	\end{table} 
	
	\section{Conclusion}
	
	In this work, we propose gated transfer network for transfer learning. By analyzing domain similarity, we argue that weighting features from pre-trained CNN models for target domains is a more effective approach. Hence, we introduce a transfer module as a feature selection mechanism with minimal extra computational cost. The novel transfer module can be interpreted as classic fine-tuning or sparse coding depending on the learned gating scores. In addition, the transfer module can be easily incorporated with other transfer learning techniques for further improvement. We evaluate our proposed GTN on six different domains and prove its effectiveness by achieving state-of-the-art results on all benchmarks. 
	
	\noindent \textbf{Acknowledgement} We gratefully acknowledge the support of NVIDIA Corporation through the donation of the Titan Xp GPUs used in this work.
	%
	%
	%
	\bibliographystyle{splncs04}
	\bibliography{egbib}
	%
	
	
	
	
\end{document}